\documentclass[runningheads]{llncs}
\usepackage{siunitx}
\usepackage{amsmath} 
\usepackage{enumitem}
 
\usepackage{eccvabbrv}
\usepackage{multirow}
\usepackage{graphicx}
\usepackage{booktabs}
\usepackage{colortbl}
\usepackage[table]{xcolor} 
\usepackage{pifont}

\usepackage[accsupp]{axessibility}  

\usepackage{hyperref}

\usepackage{orcidlink}

\begin{document}

\title{Cross-view image geo-localization with Panorama-BEV Co-Retrieval Network} 

\titlerunning{Cross-view Panorama-BEV Co-Retrieval}


\authorrunning{J.~Ye et al.}


\makeatletter
\renewcommand{\@fnsymbol}[1]{%
  \ifcase#1\or *\or † \else\@arabic{#1}\fi%
}
\makeatother

\title{Cross-view image geo-localization with Panorama-BEV Co-Retrieval Network}

\author{
    Junyan Ye\inst{1,2}\thanks{This work was partially done during the internship at Shanghai AI Laboratory.}  \and
    Zhutao Lv\inst{1} \and
    Weijia Li\inst{1}\thanks{Corresponding authors.} \and
    Jinhua Yu\inst{1} \and
    Haote Yang\inst{2} \and \\
    Huaping Zhong\inst{3} \and
    Conghui He\inst{2,3 }\textsuperscript{†}
}

\institute{
    \makebox[0.9\textwidth][c]{\inst{1} Sun Yat-Sen University, \inst{2} Shanghai AI Laboratory, \inst{3} SenseTime Research} \\
    \email{\{yejy53,lvzht5\}@mail2.sysu.edu.cn}, \email{liweij29@mail.sysu.edu.cn}, \\ \email{\{yanghaote,heconghui\}@pjlab.org.cn}, \email{zhonghuaping@sensetime.com}
}
\maketitle

\begin{abstract}

Cross-view geolocalization identifies the geographic location of street view images by matching them with a georeferenced satellite database. Significant challenges arise due to the drastic appearance and geometry differences between views. In this paper, we propose a new approach for cross-view image geo-localization, i.e.,  the Panorama-BEV Co-Retrieval Network. Specifically, by utilizing the ground plane assumption and geometric relations, we convert street view panorama images into the BEV view, reducing the gap between street panoramas and satellite imagery. In the existing retrieval of street view panorama images and satellite images, we introduce BEV and satellite image retrieval branches for collaborative retrieval. By retaining the original street view retrieval branch, we overcome the limited perception range issue of BEV representation. Our network enables comprehensive perception of both the global layout and local details around the street view capture locations. Additionally, we introduce CVGlobal, a global cross-view dataset that is closer to real-world scenarios. This dataset adopts a more realistic setup, with street view directions not aligned with satellite images. CVGlobal also includes cross-regional, cross-temporal, and street view to map retrieval tests, enabling a comprehensive evaluation of algorithm performance. Our method excels in multiple tests on common cross-view datasets such as CVUSA, CVACT, VIGOR, and our newly introduced CVGlobal, surpassing the current state-of-the-art approaches. The code and datasets can be found at \url{https://github.com/yejy53/EP-BEV}.

    
\end{abstract}


\section{Introduction}
\label{sec:intro}

Cross-view retrieval geolocalization involves matching ground images with georeferenced satellite images in a database to identify their geographical locations \cite{hu2018cvm,liu2019lending,shi2019spatial,hu2020image,Deuser_2023_ICCV,zhu2021VIGOR,sarlin2024snap,thoma2019mapping}, as shown in Fig. \ref{fig:intro} (a). 
Cross-view retrieval faces challenges due to the significant differences between satellite and ground imagery perspectives. For example, buildings appear differently in satellite view from their rooftops compared to ground perspectives of their facades. The morphological and textural characteristics of these viewpoints vary significantly. However, some elements, such as roads and crops are observable from both ground and satellite viewpoints, despite their visual differences, representing cross-view shared information \cite{zhu2023cross}. The task focuses on harnessing cross-view information to effectively align content and distributions across both perspectives.

Current cross-view retrieval methods primarily leverage deep learning techniques with CNN \cite{workman2015wide,hu2018cvm,cai2019ground,shi2020optimal,wang2021each,shi2020looking} and Transformer \cite{yang2021cross,zhu2022transgeo,zhu2023simple}  architectures to transform images from different perspectives into feature vectors . These vectors are then matched based on similarity calculations in the feature space. However, aligning the embedded feature vectors in the spatial domain remains challenging due to significant perspective differences. To mitigate this issue, some approaches use polar coordinate transformations to reduce geometric differences \cite{shi2019spatial,toker2021coming}. Specifically, instead of directly matching satellite view with street view images, satellite view images are first transformed into polar view images before being matched with street view images, as shown in Fig. \ref{fig:intro} (b). The polar transformation effectively aligns cross-view shared information, such as road orientations, achieving notable performance improvements. However, transformed polar view images still exhibit significant differences in information distribution compared to ground images. For instance, ground images often include some sky information, while polar coordinate-transformed images contain treetop information, with considerable morphological distortion. 

We observe that, in addition to converting satellite images to street view perspectives, it's also feasible to transform street view to satellite viewpoints. We transform street view panoramas into explicit Bird's Eye View (BEV) images using azimuth relationships and ground plane constraints. Compared to polar transformation, transforming street view into satellite viewpoints is more intuitive, resulting in transformed images that are more realistic and highlight cross-view shared local information near the shooting location. On the other hand, since our BEV transformation does not rely on depth and 3D structure estimation, images transformed via BEV can exhibit limited visibility and severe distortion in dense urban scenes where they are obstructed by tall structures like buildings. To tackle this challenge, we designed the Panorama-BEV Co-Retrieval Network,  which collaboratively leverages street view panoramas and BEV images for satellite image retrieval. We retain the original street view panorama to satellite retrieval branch to expand the perception range and capture more global layout features, while the BEV to satellite retrieval branch focuses on the details near the street view locations.

Current cross-view retrieval research primarily utilizes datasets like CVUSA\cite{workman2015wide} , CVACT \cite{liu2019lending}, and VIGOR\cite{zhu2021VIGOR}, with CVUSA achieving top-1 recall rates over 98\%, demonstrating the effectiveness of cross-view methods. However, a gap still exists between these datasets and real-world applications. Firstly, existing datasets mainly focus on a single country, limiting evaluations across diverse global scenes. Secondly, challenges with street views lacking metadata extend beyond unknown locations to include uncertain camera orientations and shooting times. Currently, there are few datasets with street views that have uncertain orientations; most approaches simulate random captures by rotating fixed-orientation street view images \cite{shi2020looking,zhu2022transgeo}. Additionally, there is a lack of cross-temporal retrieval task evaluations, raising questions about whether current satellite imagery can accurately locate street views captured at unknown times. Furthermore, there have been few attempts to use map data instead of satellite data in the current datasets. Map data has advantages over satellite data, such as easier accessibility and storage. To address these challenges, we introduce a global cross-view retrieval dataset named CVGlobal. This dataset includes cross-regional, cross-temporal, and street view to map retrieval tests, aiming for a comprehensive assessment of algorithmic performance.

Our main contributions can be summarized as follows:

\begin{figure}[!t]
    \centering
    \includegraphics[width=\linewidth]{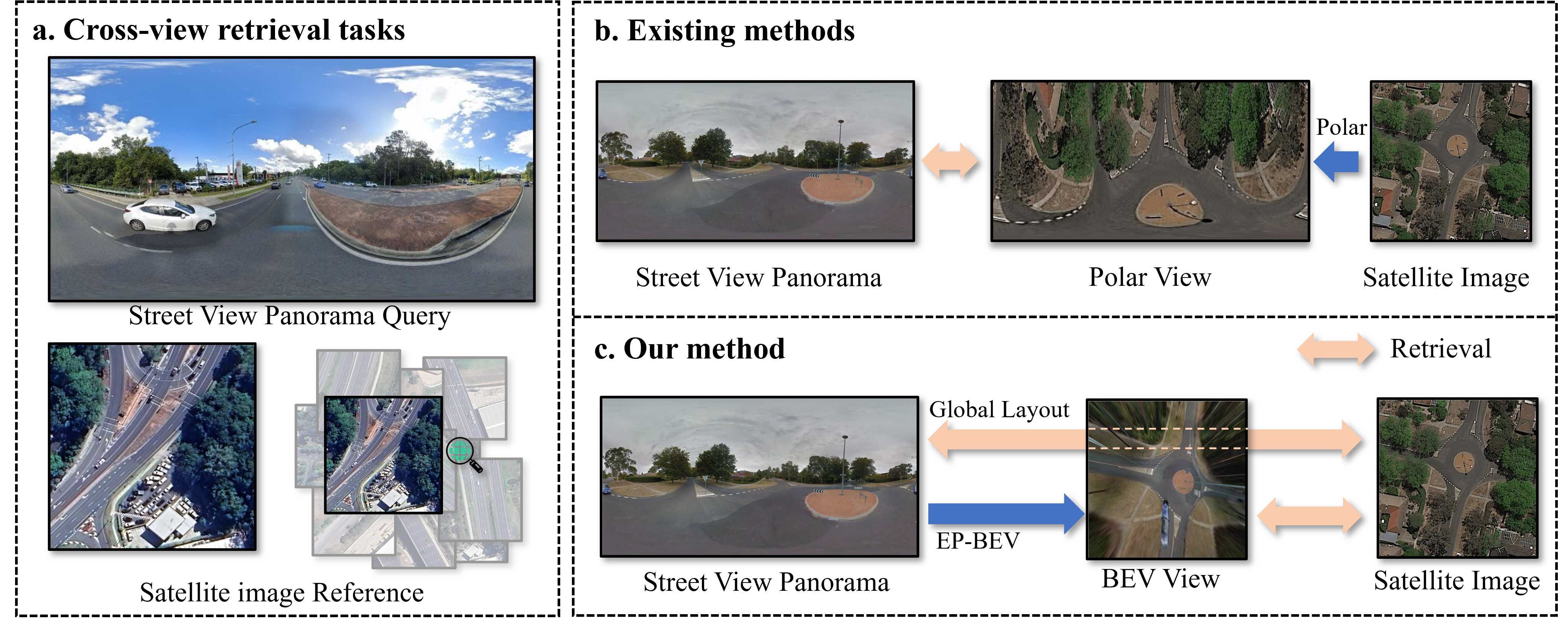}
    \caption{The goal of cross-view retrieval is to identify the georeferenced satellite image most visually similar to the street view panorama query (a). Existing methods use polar transformation to convert satellite views into Polar views and then proceed with retrieval (b). Our method employs Explicit Panoramic BEV transformation to convert street view images into the BEV perspective consistent with satellite views, while preserving the previous street view panorama to satellite retrieval path (c). }
    \label{fig:intro}
\end{figure}

\begin{itemize}[label={\textbullet}, font=\bfseries, leftmargin=*]
  \item We propose a novel transformation approach for cross-view retrieval tasks, explicitly converting street view panoramas into BEV views, effectively bridging the gap between street and satellite perspectives. By designing the Panorama-BEV Co-Retrieval Network, we facilitate collaborative satellite retrieval with street view panoramas and BEV images, surpassing BEV's perceptual limits to fully perceive global layouts and local details.
  \item We introduce CVGlobal, a global cross-view retrieval dataset that is closer to real-world application scenarios. The dataset features indeterminate street view orientations and supports evaluations of cross-regional, cross-temporal, and street view to map retrieval tasks.
  \item Our method has been extensively evaluated across multiple datasets and outperforms the current state-of-the-art approaches. In challenging cross-regional tasks such as VIGOR-cross and from CVUSA to CVACT, our method improves the top-1 recall rate, demonstrating the generalization capability.
\end{itemize}

\section{Related Work}

\subsection{Cross-view retrieval}
Cross-view image retrieval methods use ground images as queries and all patches in a satellite image database as references for geolocation. Early retrieval efforts relied on manual features to match images across the two domains \cite{bansal2011geo,toker2021coming}. With the advent of deep learning algorithms, methods have evolved to embed images into global feature descriptors for retrieval \cite{lin2013cross,workman2015wide,vo2016localizing,tian2017cross,zhai2017predicting}. Deuser et al. \cite{Deuser_2023_ICCV} employed the infoNCE loss combined with global hard negative mining, achieving state-of-the-art results. To mitigate the significant differences between satellite and ground images, many studies have improved retrieval accuracy through polar coordinate transformation algorithms \cite{shi2019spatial,shi2022accurate,yang2021cross,zhu2022transgeo}. Polar coordinate transformation, which relies on orientation relationships for direct conversion, introduces certain distortions when converting satellite images to street views. Toker et al. \cite{toker2021coming} leveraged GANs \cite{goodfellow2020generative} to learn to eliminate these distortions. 

Current algorithms perform well in top-5 and top-10 recall rates but struggle with low top-1 recall due to the challenge of distinguishing similar images in dense scenes when embedding street and satellite images directly. Our method enhances distinguishability by employing transformed BEV images for retrieval, incorporating more features near the shooting location.

\subsection{BEV transformation}
Transforming ground images into Bird's Eye View (BEV) representations is a key method for tasks such as autonomous driving and localization \cite{peng2023bevsegformer,reiher2020sim2real,pan2020cross,sgbev}. However, current methods based on BEV have a high demand for camera intrinsic and extrinsic parameters. 
OrienterNet \cite{sarlin2023orienternet}  achieves precise localization with known approximate GPS positions by estimating camera parameters and scene depth for BEV feature mapping. Boosting \cite{shi2023boosting} explores BEV feature-level projection based on geometric methods, yet converting tens of thousands of satellite images in a database to BEV feature representations, instead of efficient vector representations, remains a significant cost issue for retrieval tasks. Wang et al.\cite{wang2024fine} conducted explicit image transformations for cross-view localization tasks, achieving good results. However, this requires knowing the corresponding satellite image for a ground image, which is a subsequent task after retrieval.

Our Explicit Panoramic BEV Transformation utilizes geometric relationships and the ground plane assumption, starting from a predefined BEV plane to inversely calculate the panorama's indices, achieving explicit BEV transformation without the need for intrinsic or depth estimation. Unlike many localization approaches mapping street views to BEV feature representations with higher computational costs, our method converts BEV into an image representation, enabling direct feature embedding and efficient searching.

\subsection{Cross-view Datasets}
Several cross-view geolocation datasets have been introduced, including CVUSA \cite{workman2015wide}, CVACT \cite{liu2019lending}, Vo\cite{vo2016localizing}, Universities-1652 \cite{zheng2020university}, and VIGOR \cite{zhu2021VIGOR}. The CVUSA dataset features 355,332 ground-to-satellite image pairs from the United States, while CVACT has a similar training/validation volume and a larger test set, CVACT-test. CVUSA and CVACT are the most commonly used cross-view retrieval datasets, employing a one-to-one retrieval setup. The VIGOR dataset includes data from multiple cities and it evaluates the model's transferability across different geographic regions. In this dataset, street panoramas and satellite images are not centrally aligned. Multiple street-view images cover the same satellite image area, with overlapping regions between different satellite images. Vo's research collected paired images from 11 U.S. cities, combining street views with Google Maps satellite photos. Universities-1652 extends the dataset by incorporating drone imagery in addition to street and satellite images. 

Existing cross-view datasets provide comprehensive evaluations of cross-view retrieval algorithms across multiple dimensions and tasks. However, the existing datasets still fall short of real-world application scenarios. These include the need for data from more diverse cities worldwide, varying street view orientations, considering street view image retrieval at different times, or using map data instead of satellite images for retrieval. To tackle these issues, we introduce CVGlobal, a global cross-view retrieval dataset. It features street views with non-fixed orientations and includes cross-regional, cross-temporal, and street-to-map retrieval tests, aiming for a comprehensive evaluation of algorithm performance.

\begin{figure}[!t]
    \centering
    \includegraphics[width=\linewidth]{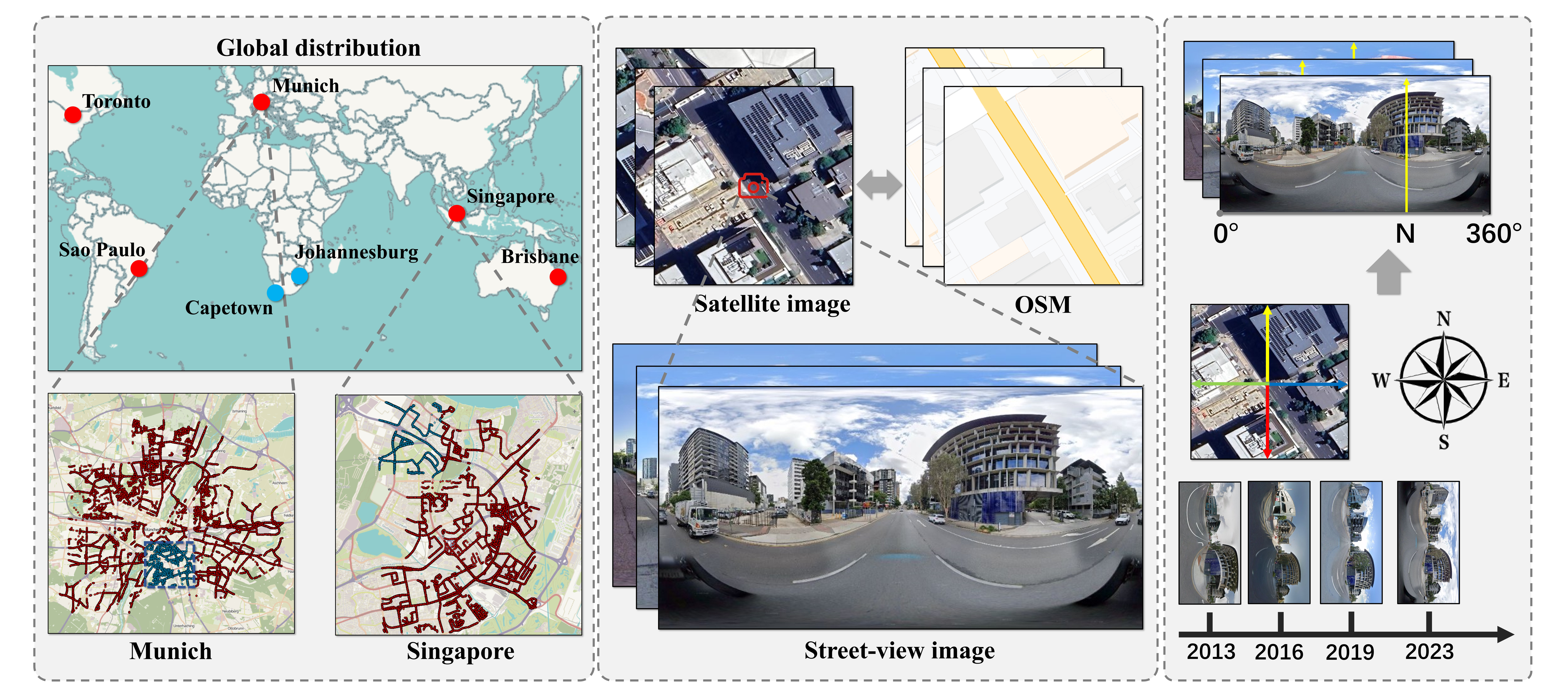}
    \caption{The cross-view retrieval dataset CVGlobal encompasses data from various distinct style cities around the world, with red sample points representing training data and blue points indicating regional testing data (a). Since street views are captured by car-mounted cameras, they are usually centered on the road, and the north direction is not fixed (b). Additionally, CVGlobal introduces new tasks such as cross-temporal evaluation (c) and street view to map evaluation (d).}
    \label{fig:dataset}
\end{figure}
\section{Dataset}
\label{sec:Dataset}

\subsection{Dataset collection}

We downloaded 134,233 street view images from seven cities globally in 2023 using Google Street View Download 360\footnote{\url{https://svd360.istreetview.com/}}, including Munich, Toronto, Singapore, São Paulo, Brisbane, Cape Town, and Johannesburg, with an average distance of 50 m between images. Additionally, street views from Brisbane for 2013, 2016, and 2019 were collected to evaluate the algorithm's cross-temporal retrieval capabilities.  Using the Google Maps Static API\footnote{\url{https://developers.google.com/maps/documentation/maps-static/}}, we acquired corresponding satellite images and map data based on the latitude and longitude of the street views. The satellite images were at a size of \(512 \times 512\), covering a spatial range of about \(70\text{m} \times 70 text{m}\). Map data and satellite imagery share the same coverage area and resolution. 

\begin{table}[!t]
    \centering
    \scriptsize
    \caption{Comparison of the proposed CVGlobal dataset with existing datasets in cross-view retrieval.}
    \begin{tabular}{l|c|c|c|c|c|c}
        \toprule
         & Vo \cite{vo2016localizing} & Uni.-1652\footnotemark[1] \cite{zheng2020university} & CVUSA \cite{workman2015wide} & CVACT \cite{liu2019lending} & VIGOR \cite{zhu2021VIGOR} & CVGlobal \\
        \midrule
        Satellite images & $\sim$ 450,000 & 50,218 & 44,416 & 128,334 & 90,618 & 134,233 \\
        Query images & $\sim$ 450,000 & 41,135 & 44,416 & 128,334 & 105,214 & 134,233 \\
        Full panorama & \ding{55} & \ding{55} & \ding{51} & \ding{51} & \ding{51} & \ding{51} \\
        Center aligned & \ding{51} & \ding{55} & \ding{51} & \ding{51} & \ding{55} & \ding{51} \\
        \rowcolor{lightgray!40} Global scale & \ding{55} & \ding{51} & \ding{55} & \ding{55} & \ding{55} & \ding{51} \\
        \rowcolor{lightgray!40} Unfixed Orientation & \ding{51} & \ding{51} & \ding{55} & \ding{55} & \ding{55} & \ding{51} \\
        \rowcolor{lightgray!40} Multiple time & \ding{55} & \ding{55} & \ding{55} & \ding{55} & \ding{55} & \ding{51} \\
        \rowcolor{lightgray!40} Map supplement & \ding{55} & \ding{55} & \ding{55} & \ding{55 }& \ding{55} & \ding{51} \\
        \bottomrule
    \end{tabular}
    \label{tab:datasets}
    \raggedright\footnotemark[1]{Uni.-1652 refers to the Universities-1652 dataset.}
\end{table}

\subsection{Dataset comparison}

Table \ref{tab:datasets} showcases a comparison between our dataset and previous benchmarks, illustrating that our dataset is closer to real-world scenarios with more potential application. Covering cities with a wide range of styles allows for an effective assessment of the algorithm's robustness across various scenarios. Additionally, the orientation of street views is not fixed. The dataset also includes street view data from Brisbane over multiple years, allowing for the evaluation of cross-temporal retrieval tasks. We use street view data from past years as queries and current satellite imagery as references. Retrieving images from different time periods represents a novel endeavor. Moreover, we've gathered map data slices aligned with satellite imagery, setting up street view to map slice retrieval tasks to probe their utility in cross-view retrieval. We use street view images as queries and rasterized map data slices as references. Compared to the high capture and storage costs of high-resolution satellite images, map data is easier to acquire and store. However, map data lacks the texture information present in satellite views, retaining only partial shape information. Particularly in underdeveloped areas, where statistical data is sparse and updates are slow, map data contains very little useful information, posing challenges to the task.

\subsection{Evaluation schemes} 
We selected street view satellite data from selected areas of Munich, Toronto, Singapore, São Paulo, and Brisbane from the year 2023 as our training set. Similar to CVUSA, we used data randomly divided from the same regions as the training set for our validation set. To address real-world application scenarios, we designed multiple experimental evaluation schemes: 

\textbf{Cross-regional retrieval.} 
Our cross-regional tests are of two types: one within different areas of the training cities, as illustrated by the blue areas in Fig. \ref{fig:dataset}, and the second using Cape Town and Johannesburg in Africa as test sets, increasing the task's difficulty. During testing, the satellite image database corresponding to the query images only includes the regional test set.

\textbf{Cross-temporal retrieval.} As mentioned earlier, our model includes Brisbane's training data from 2023, then tests its cross-temporal performance. We use street view images from Brisbane from the years 2013, 2016, and 2019 as queries, and satellite images from the corresponding locations in 2023 as the database to investigate whether the algorithm performs well across different years. We also combined data from these three years as input to investigate changes in the algorithm's performance.

\textbf{Street view to map retrieval.} We replaced the corresponding satellite images with map data for retraining and testing. We employed evaluations consistent with satellite imagery to explore the application potential of map data.

\section{Methods}
\label{sec:methods}

\subsection{Overview} 
In cross-view retrieval task, the goal is to identify the most similar satellite imagery in a database based on the visual features of an input street view panorama query, thereby achieving geolocation of street view data. The primary challenge of this task lies in the significant perspective difference between street and satellite imagery. We address this by employing an explicit panoramic BEV transformation to bridge the gap between the two domains, highlighting cross-view information. Furthermore, to overcome the limited observation range of the BEV's fidelity mapping, we additionally utilize street view panorama branches to access a broader range of global observations. 

\begin{figure}[!t]
    \centering
    \includegraphics[width=\linewidth]{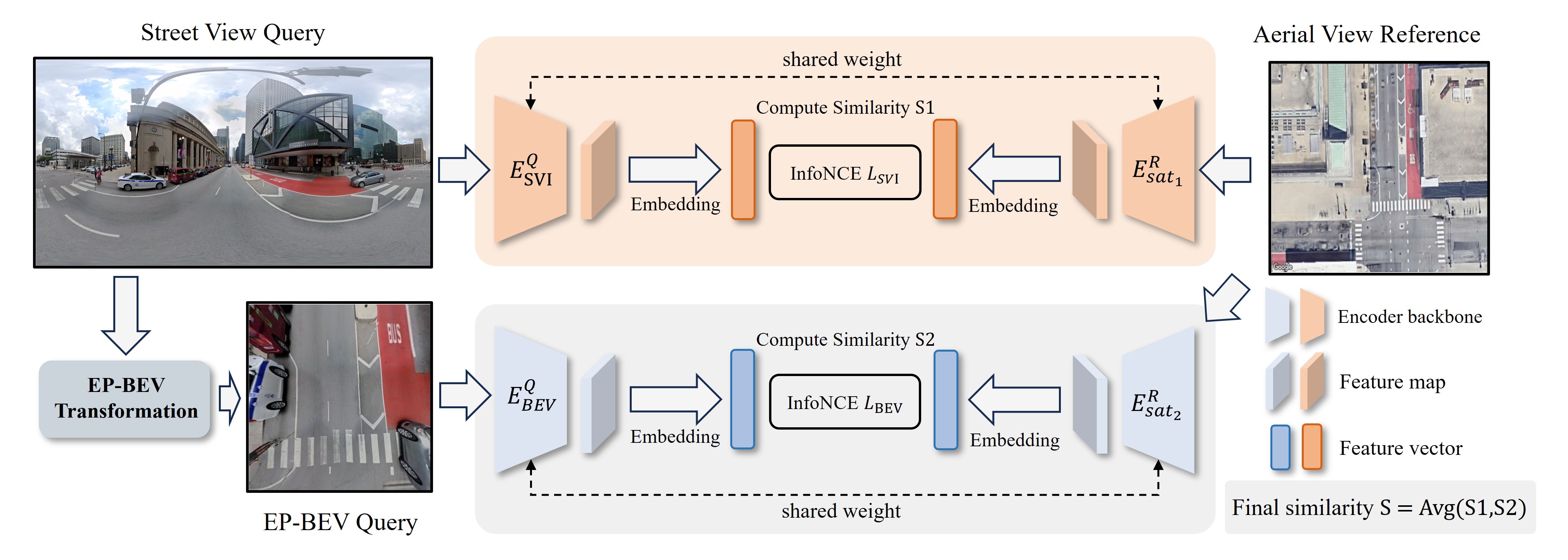}
    \caption{Schematic of the Panorama-BEV Co-Retrieval Network. Street view images and their images transformed via EP-BEV serve as query inputs, with satellite images as reference inputs. The network comprises a street view branch focusing on matching global layout information with satellite images and a BEV branch emphasizing detailed feature matching between the nearby street view area and the satellite perspective. }
    \label{fig:pipeline}
\end{figure}

As illustrated in Fig. \ref{fig:pipeline}, this paper introduces a novel cross-view retrieval method, Panorama-BEV Co-Retrieval Network. In the BEV branch, street view images are transformed into the satellite perspective through EP-BEV transformation for retrieval (see Section \ref{sec:4.2}). Meanwhile, the street view panorama branch directly uses panoramas to search for satellite images. We achieve collaborative retrieval by simultaneously utilizing street view panoramas and BEV images (see Section \ref{sec:4.3}). We trained two models for the street view and BEV branches using the same contrastive image retrieval objective. During testing, the network will simultaneously apply both branches to retrieve the matching images for a given street view query, and the final decision will be made by combining the similarity scores from both branches.

\subsection{Explicit panoramic BEV transformation}\label{sec:4.2}
 
\begin{figure}[!t]
    \centering
    \includegraphics[width=0.8\linewidth]{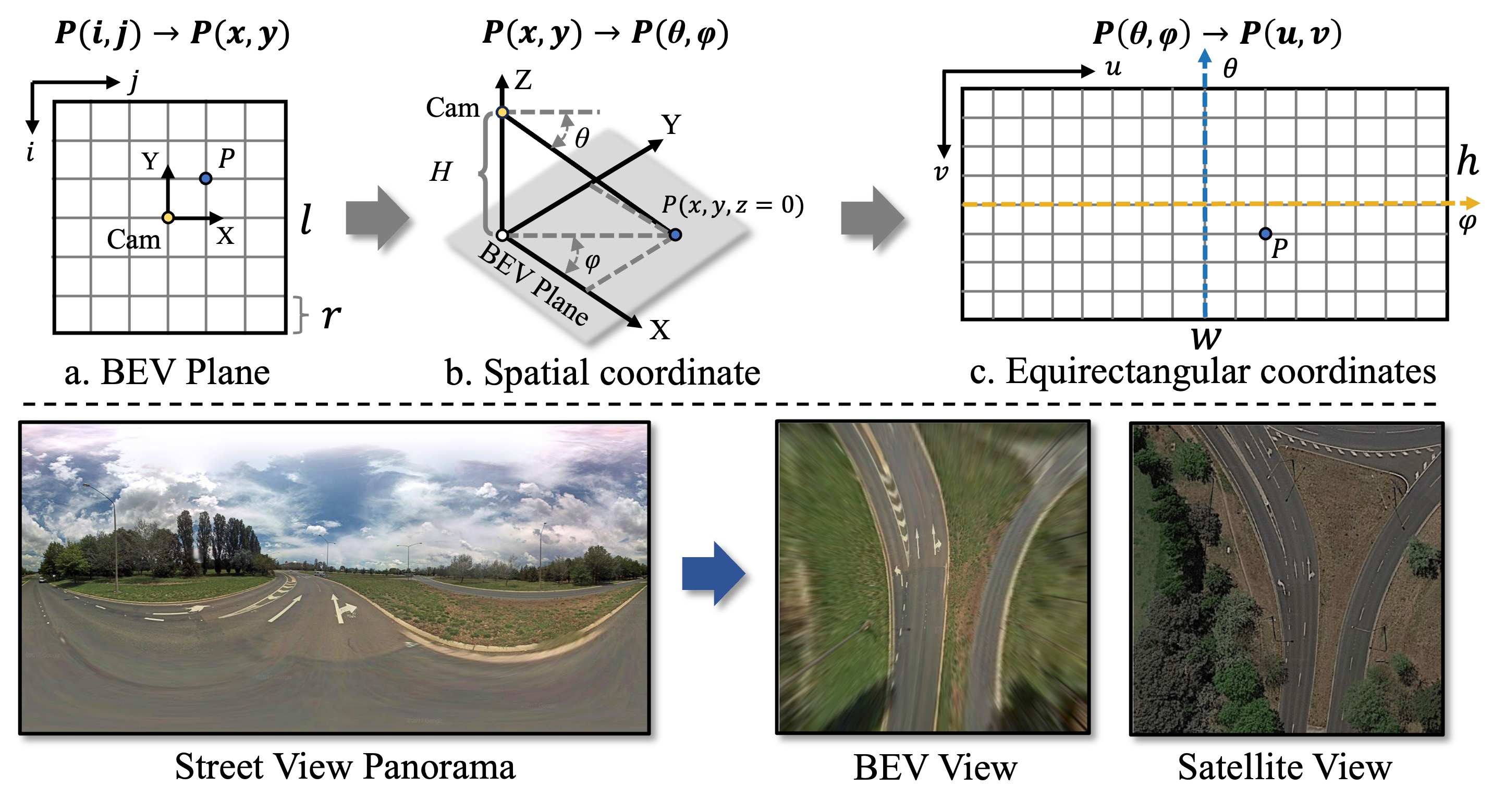}
    \caption{Schematic of the Explicit Panoramic BEV Transformation, with the upper part showing specific transformation details and the lower part displaying the results of the BEV conversion.}
   \label{fig:search}
\end{figure}

Traditional Bird's Eye View (BEV) transformation processes rely on accurate estimation of depth information and camera parameters. In contrast, our proposed method utilizes a geometric back-projection process based on the ground plane assumption, which directly calculates the corresponding positions of points on the BEV plane in the panorama (as shown in Fig. \ref{fig:search}).

Given our objective to transform street view images into BEV views spatially aligned with satellite imagery, we first define a predetermined BEV plane aligned with the satellite perspective (as shown in Fig. \ref{fig:search} (a)), assuming the camera is located at the center of this BEV plane. Next, utilizing the grid relationship of the plane, \(i, j\), we can determine the coordinates of the required mapping point \(P(x, y, z=0)\) (see Eq. \ref{eq:mapping_point}). By setting the camera height to \(H\) and positioning the camera at \(Cam(0,0,H)\), we establish the spatial coordinate system (as illustrated in Fig. \ref{fig:search} (b)). Using geometric relationships, we can calculate the corresponding pitch angle \(\theta\) and azimuth angle \(\varphi\) (see Eq. \ref{eq:angles}). With the equirectangular cylindrical projection characteristic of panoramic images, we can use \(\theta\) and \(\varphi\) to calculate the respective row and column numbers \(v, u\) (see Eq. \ref{eq:projection}). By mapping the index relationship between \(i, j\) and \(v, u\), we achieve the image transformation from the street view perspective to the BEV perspective. More details will be found in the supplementary material.

\begin{equation}\label{eq:mapping_point}
x = \left( j - \frac{l}{2} \right) \times r
, \qquad
y = \left( \frac{l}{2} - i \right) \times r
\end{equation}
\vspace{-4pt}
\begin{equation}\label{eq:angles}
\theta = - \arctan\left(\frac{H}{\sqrt{x^2 + y^2}}\right), \qquad
\varphi = \arctan2(y, x)
\end{equation}
\vspace{-4pt}
\begin{equation}\label{eq:projection}
\textit{v} = \left( \frac{\pi}{2} - \theta \right) \times \frac{h}{\pi}, \qquad
\textit{u} = \left(\frac{\varphi+\pi}{2\pi}\right) \times w
\end{equation}

In the formulas, \(y\) and \(x\) represent the coordinates in three-dimensional space, which are calculated through the row number \(i\) and column number \(j\) on the BEV plane, the edge length of the BEV plane is  \(l\), and its resolution is \(r\). The camera height is set to \(H\), the pitch angle \(\theta\) refers to the angle between the line connecting the camera to point \(P\) and the camera plane and the azimuth angle \(\varphi\) is the angle with respect to the positive direction of the x-axis. \(v\) and \(u\)  are the row and column numbers of the panoramic image, respectively, while \(h\) and \(w\) are the height and width of the panoramic image.

Through explicit panoramic BEV transformation, we project street view images into a bird's-eye view without needing depth estimation or camera parameters. Although transformed images lose information on tall objects, like building facades, this information is unique to the ground perspective and not visible from the satellite view. Explicit panoramic BEV transformation effectively minimizes the correspondence gap between the two domains, highlighting information on objects observable in both satellite and street views.

\subsection{Dual-branch cross-view image retrieval}\label{sec:4.3}

Our method employs a dual-branch structure to accomplish the collaborative retrieval (co-retrieval) task of street view panoramas and BEV. For the street view retrieval branch, we embed street and satellite images as feature vectors through encoders and optimize using InfoNCE loss $L_{Pan}$. This branch directly utilizes the original street view inputs, covering a wider observation range and aligning with satellite images in a more global layout distribution. In the BEV retrieval branch, following the transformation described in Section \ref{sec:4.2}, we first convert street views into BEV views to perform detailed feature alignment under the satellite perspective, optimizing with InfoNCE loss $L_{BEV}$. This branch, using the transformed EP-BEV inputs, emphasizes cross-view information near the street view, aiding in distinguishing between similar satellite images. Since the optimization directions of the different branches are not the same, the two branches are trained separately to obtain the final models.

\begin{equation}
L = -\log \left( \frac{\exp(Q \cdot R^+ / \tau)}{\sum_{j=1}^{N} \exp(Q \cdot R_{j} / \tau)} \right)
\end{equation}

In this formula, $Q$ represents the query images, including the street view panorama query image $Q_{Pan}$ and the BEV query image $Q_{BEV}$. $R^+$ represent the positive reference images that are geographically consistent with the query images. $R_{j}$ denote the negative samples and $\tau$ is a temperature parameter. During inference, the street view branch computes similarity S1, and the EP-BEV branch computes similarity S2. The sum of S1 and S2 determines the final retrieval result. To save memory, only the top-ranked results from the street view branch need to be used for similarity merging.

\section{Experiments}
\label{sec:experiments}

\subsection{Dataset and Evaluation Protocol}

Following common experimental  setting \cite{shi2019spatial,shi2020looking,zhu2022transgeo,zhu2023simple,Deuser_2023_ICCV,yang2021cross}, we conducted extensive experimental evaluations on three widely used datasets: CVUSA \cite{workman2015wide}, CVACT \cite{liu2019lending}, VIGOR \cite{zhu2021VIGOR} and our proposed CVGlobal dataset, to validate the effectiveness of our model. We utilized the metric of top-k image recall rate to assess model performance. Specifically, given a street view panorama query, if the corresponding closest satellite image is within the top k retrieved images, then the query is considered "successfully retrieved." The percentage of query images that have been correctly localized is referred to as R@K.

\subsection{Implementation Details}
We employ the ConvNeXt-B \cite{liu2022convnet} as the backbone network for encoding both ground and satellite images in the two retrieval stages, utilizing the AdamW optimizer with a learning rate set to \(1.0 \times 10^{-3}\). The temperature parameter $\tau$ is a learnable parameter \cite{Deuser_2023_ICCV}. Our training spans 40 epochs, with a batch size of 128. Following the setup by Yu et al. \cite{shi2023boosting}, we set the camera shooting height to 1.5 m, and the BEV plane range is aligned with the satellite imagery range of CVACT, with the image size \( l \) set to \( 512 \times 512 \) and a pixel resolution \( r \) of 14cm. Since CVUSA's panoramic images are not of regular size, ground images need to be padded to standard dimensions. All comparison methods were implemented according to their publicly available source code settings.

\begin{table}[!t]
    \scriptsize
    \centering
    \caption{Quantitative comparison between our approach and state-of-the-art approaches on CVUSA and CVACT. $\dag$ denotes methods that use polar transformation.}
    \setlength{\tabcolsep}{1.5pt}
    \begin{tabular}{@{}l|cccc|cccc|cccc@{}}
        \toprule
        \multirow{2}{*}{Methods} & \multicolumn{4}{c|}{CVUSA} & \multicolumn{4}{c|}{CVACT Val} & \multicolumn{4}{c}{CVACT Test} \\
         & R@1 & R@5 & R@10 & R@1\% & R@1 & R@5 & R@10 & R@1\% &  R@1 & R@5 & R@10 & R@1\% \\
        \midrule
        SAFA$^{\dag}$\cite{shi2019spatial}  & 89.84 & 96.93 & 98.14 & 99.64 & 81.03 & 92.80 & 94.84 & - & - & - & - & - \\
        LPN\cite{wang2021each} & 85.79 & 95.38 & 96.98 & 99.41 & 79.99 & 90.63 & 92.56 & - & - & - & - & - \\
        LPN$^{\dag}$\cite{wang2021each} & 92.83 & 98.00 & 98.85 & 99.78 & 83.66 & 94.14 & 95.92 & 98.41 & - & - & - & - \\
        DSM\cite{shi2020looking} & 91.96 & 97.50 & 98.54 & 99.67 & 82.49 & 92.44 & 93.99 & 97.32 & - & - & - & - \\
        TransGeo \cite{zhu2022transgeo} & 94.08 & 98.36 & 99.04 & 99.77 & 84.95 & 94.14 & 95.78 & 98.37 & - & - & - & - \\
        GeoDTR\cite{zhang2023cross} & 93.76 & 98.47 & 99.22 & 99.85 & 85.43 & 94.81 & 96.11 & 98.26 & 62.96 & 87.35 & 90.70 & 98.61 \\
        GeoDTR$^{\dag}$\cite{zhang2023cross} & 95.43 & 98.86 & 99.34 & 99.86 & 86.21 & 95.44 & 96.72 & 98.77 & 64.52 & 88.59 & 91.96 & 98.74 \\
        SAIG-D\cite{zhu2023simple} & 96.08 & 98.72 & 99.22& 99.86 & 89.21 & 96.07 & 97.04 & 98.74 & 67.49 & 89.39 & 92.30 & 96.80 \\
        Samp4G\cite{Deuser_2023_ICCV} & 98.68 & 99.68 & \textbf{99.78} & \textbf{99.87} & 90.81 & 96.74 & 97.48 & 98.77 & 71.51 & 92.42 & 94.45 & 98.70 \\
        Ours & \textbf{98.71} & \textbf{99.70} & \textbf{99.78} & 99.86 & \textbf{91.90} & \textbf{97.23} & \textbf{97.84} & \textbf{98.84} & \textbf{73.68} & \textbf{93.53} & \textbf{95.11} & \textbf{98.81} \\
        \bottomrule
    \end{tabular}
    \label{tab: result1}
\end{table}

\subsection{Evaluation results on existing datasets}

\begin{table}[!t]
    \scriptsize
    \setlength{\tabcolsep}{3pt}
    \centering
    \caption{Quantitative comparison between our approach and the current state-of-the-art on VIGOR. ${\dag}$ denotes methods that use polar transformation on the satellite input image.}
    \begin{tabular}{l|cccc|cccc}
        \toprule
        \multirow{2}{*}{Methods} & \multicolumn{4}{c|}{Same-area} & \multicolumn{4}{c}{Cross-area} \\
         & R@1 & R@5 & R@10 & R@1\% & R@1 & R@5 & R@10 & R@1\%\\
        \midrule
        SAFA$^{\dag}$\cite{shi2019spatial} & 33.93 & 58.42 & 68.12 & 98.24 & 8.20 & 19.59 & 26.36 & 77.61 \\
        TransGeo\cite{zhu2022transgeo} & 61.48 & 87.54 & 91.88 & 99.56 & 18.99 & 38.24 & 46.91 & 88.94\\
        SAIG-D\cite{zhu2023simple}& 65.23 & 88.08 & - & 99.68 & 33.05 & 55.94 & - & 94.64\\
        Samp4G\cite{Deuser_2023_ICCV} & 77.86 & 95.66 & 97.21 & 99.61 & 61.70 & 83.50 &  88.00 & 98.17 \\
        Ours & \textbf{82.18} & \textbf{97.10} & \textbf{98.17} & \textbf{99.70} & \textbf{72.19} & \textbf{88.68} & \textbf{91.68} & \textbf{98.56}\\
        \bottomrule
    \end{tabular}
    \label{tab: result2}
\end{table}

\begin{table}
    \scriptsize
    \setlength{\tabcolsep}{3pt}
    \centering
    \caption{Cross-dataset generalization capability test. Models are trained on the CVUSA training splits and tested on the CVACT validation and test splits. $\dag$ denotes approaches that used the polar transformation. Only SVI denotes retrieval exclusively with the street view branch, Only BEV indicates retrieval solely via the BEV branch.}
    \begin{tabular}{l|cccc|cccc}
    \toprule
    \multirow{2}{*}{Methods} & \multicolumn{4}{c|}{CVUSA $\rightarrow$ CVACT-Val} & \multicolumn{4}{c}{CVUSA $\rightarrow$ CVACT-Test} \\
     & R@1 & R@5 & R@10 & R@1\% & R@1 & R@5 & R@10 & R@1\% \\
    \midrule
    L2LTR\cite{yang2021cross}  & 47.55 & 70.58 & - & 91.39 & - & - & - & - \\
    L2LTR$^{\dag}$\cite{yang2021cross} & 52.58 & 75.81 & 77.39 & 93.51 & - & - & - & - \\
    GeoDTR\cite{zhang2023cross}  & 47.79 & 70.52 & - & 92.20 & 11.24 & 18.69 & 23.67 & 72.09 \\
    GeoDTR$^{\dag}$\cite{zhang2023cross} & 53.16 & 75.62 & 81.90 & 93.80 & 22.09 & 32.22& 39.59 & 85.53 \\
    Samp4G\cite{Deuser_2023_ICCV} & 56.62 & 77.79 & 87.02 & 94.69 & 27.78 & 52.08 &60.33 & 94.88 \\
    Ours (Only SVI) & 54.17 & 76.55 & 86.99 & 94.23 & 26.11 & 50.18 & 60.34 & 94.65 \\
    Ours (Only BEV) &61.92& 81.33& 85.81& 93.88  & 33.51 & 61.25 & 68.97& 94.18\\
    Ours  &\textbf{67.79}& \textbf{84.06} & \textbf{87.96} & \textbf{95.05}& \textbf{44.10} & \textbf{70.68} & \textbf{75.86} & \textbf{95.31}\\
    \bottomrule
    \end{tabular}
    \label{tab:result3}
\end{table}

\begin{figure}[!t]
    \centering
    \includegraphics[width=\linewidth]{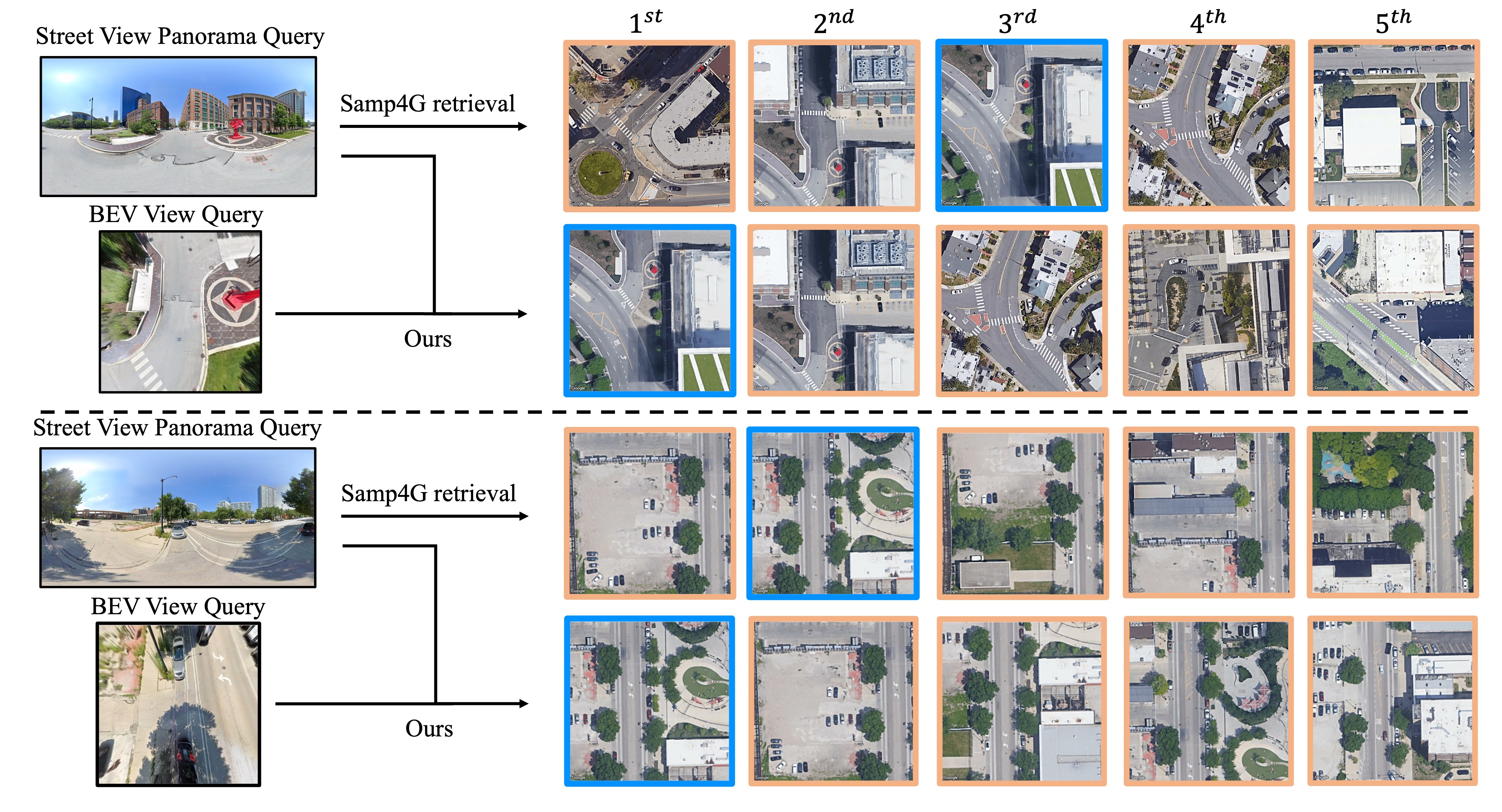}
    \caption{On the VIGOR dataset, we compare our method with Samp4G's retrieval results, using blue and orange boxes to represent correct and incorrect retrievals, respectively.}
    \label{fig:result}
\end{figure}

\textbf{Cross-view Image retrieval.} Our method performed optimally on the classic CVUSA and CVACT datasets. On the more challenging VIGOR dataset, our method increased the top-1 recall by 4.32\% in the Same-area task and by 10.49\% in the Cross-area task, indicating our method's effectiveness for difficult tasks. As visualized in Fig. \ref{fig:result} , we observe a large number of similar images within the VIGOR dataset. This phenomenon allows other methods to perform well in terms of top-10 or top-5 recall rates but show lower performance in top-1 recall. In contrast, the newly added BEV branch in our method substantially enhances the ability to distinguish details in each image and emphasizes cross-view information near the shooting location, thereby significantly improving the capability to differentiate between similar images.

\textbf{Generalisation Capabilities.}
Following the experimental setups in Samp4G \cite{Deuser_2023_ICCV} and L2LTR \cite{yang2021cross}, we evaluated the generalization capability of the algorithm, as shown in Table \ref{tab:result3}. Compared to the settings of VIGOR-cross, CVUSA and CVACT present more challenging tasks due to significant differences in resolution, satellite image size, and the scope of street view imaging. As shown in Table \ref{tab:result3}, our method scores highly on CVACT, outperforming the current state-of-the-art method by over 10\%, demonstrating strong generalizability. 
Both L2LTR\cite{yang2021cross} and GeoDTR\cite{zhang2023cross} saw significant generalization improvements with polar transformation. Our method, using only transformed BEV images, also significantly outperformed street-view-only approaches. This is attributed to the embedding of cross-view information like roads through polar and EP-BEV transformations. Direct street-to-satellite retrieval, however, is more affected by variations in observation range and resolution across datasets.

\begin{table}[!t]
    \scriptsize
    \centering
    \caption{Ablation studies for the image retrieval on the VIGOR dataset. Only SVI denotes retrieval exclusively with the street view branch, Only BEV indicates retrieval solely via the BEV branch, and Ours refers to the combination of SVI and BEV, representing our Panorama-BEV Co-Retrieval Network.}
    \setlength{\tabcolsep}{3pt}
    \begin{tabular}{l|cccc|cccc}
        \toprule
        \multirow{2}{*}{Methods} & \multicolumn{4}{c|}{Same-area} & \multicolumn{4}{c}{Cross-area} \\
         & R@1 & R@5 & R@10 & R@1\% & R@1 & R@5 & R@10 & R@1\%\\
        \midrule
        Only SVI & 76.39 & 94.76 & 96.66 & 99.58 & 58.93 & 81.28 & 86.21 & 98.01 \\
        Only BEV & 65.47 & 87.26 & 90.84 & 98.44 & 44.01 & 64.87 & 71.29 & 93.12 \\
        Ours & \textbf{82.18} & \textbf{97.10} & \textbf{98.17} & \textbf{99.70} & \textbf{72.19} & \textbf{88.68} & \textbf{91.68} & \textbf{98.56}\\
       
        \bottomrule
    \end{tabular}
    \label{result2}
\end{table}

\textbf{Ablation experiment.} We conducted ablation experiments on the VIGOR dataset to explore the effectiveness of our algorithm, as shown in Table \ref{result2}. We compared the results of retrieval using each of the two branches independently with the outcomes of our collaborative retrieval that simultaneously employs both branches. The results show that using both street view and BEV queries simultaneously significantly improves recall compared to previously using only street view. This also indicates that using only the BEV branch is not ideal due to the limited observation range compared to satellite images, resulting in a significant loss of global layout information.

\subsection{Evaluation results on the proposed CVGlobal dataset}

\begin{table}[!t]
    \centering
    \scriptsize
    \setlength{\tabcolsep}{1.5pt}
    \caption{Quantitative comparison between our approach and the current state-of-the-art on CVGlobal Cross-regional image retrieval.}
    \begin{tabular}{l|cccc|cccc|cccc}
    \toprule
        \multirow{2}{*}{Methods} & \multicolumn{4}{c|}{Regional val} & \multicolumn{4}{c|}{Regional test} & \multicolumn{4}{c}{Cross-continent test} \\
         & R@1 & R@5 & R@10 & R@1\% & R@1 & R@5 & R@10 & R@1\% & R@1 & R@5 & R@10 & R@1\% \\
        \midrule
        GeoDTR\cite{zhang2023cross} & 47.42 & 68.43 & 78.11 & 99.05 & 25.43 & 42.07 & 51.66 & 84.74 & 4.58 & 9.32 & 13.15 & 52.53 \\
        SAIG-D\cite{zhu2023simple} & 71.92 & 92.83 & 96.05 & 99.81 & 34.72 & 61.53 & 71.08 & 91.47 & 8.66 & 20.50 & 27.28 & 69.96 \\
        Samp4G\cite{Deuser_2023_ICCV} & 97.20 & 99.43 & 99.69 & \textbf{99.93} & 84.66 & 92.66 & 94.42 & 98.10 & 51.21 & 70.95 & 76.02 & 92.58 \\
        Ours & \textbf{97.78} & \textbf{99.63} & \textbf{99.79} & \textbf{99.93} & \textbf{85.65} & \textbf{92.91} & \textbf{94.77} & \textbf{98.21} & \textbf{60.25} & \textbf{75.01} & \textbf{79.51} & \textbf{93.51} \\
        \bottomrule
    \end{tabular}
    \label{tab:reginal}
\end{table}

\begin{table}[!t]
    \centering
    \scriptsize
    \setlength{\tabcolsep}{2pt}
    \caption{Quantitative comparison between our approach and the current state-of-the-art on CVGlobal Cross-temporal image retrieval.}
    \begin{tabular}{l|ccc|ccc|ccc|ccc}
        \toprule
        \multirow{2}{*}{Methods} & \multicolumn{3}{c|}{2013} & \multicolumn{3}{c|}{2016} & \multicolumn{3}{c|}{2019} & \multicolumn{3}{c}{Mixing} \\
         & R@1 & R@5 & R@10 & R@1 & R@5 & R@10 & R@1 & R@5 & R@10 & R@1 & R@5 & R@10 \\
        \midrule
        GeoDTR\cite{zhang2023cross} & 10.05 & 19.33 & 25.93 & 12.10 & 22.74 & 30.46 & 13.76 & 24.94 & 32.82 & 7.10 & 13.81 & 19.22 \\
        SAIG-D\cite{zhu2023simple} & 13.72 & 30.83 & 40.46 & 18.77 & 39.23 & 49.31 & 22.52 & 46.26 & 56.56 & 12.52 & 28.59 & 36.92 \\
        Samp4G\cite{Deuser_2023_ICCV} & 68.59 & 84.68 & 88.35 & 82.86 & 93.11 & 94.88 & 86.61 & 96.77 & 97.87 & 73.33 & 88.56 & 91.30 \\
        Ours & \textbf{74.52} & \textbf{89.02} & \textbf{91.89} & \textbf{86.44} & \textbf{94.96} & \textbf{96.35} & \textbf{88.74} & \textbf{97.56} & \textbf{98.45} & \textbf{78.18} & \textbf{91.50} & \textbf{93.52} \\
        \bottomrule
    \end{tabular}
    \label{tab:time}
\end{table}

\begin{table}[!h]
    \centering
    \scriptsize
    \setlength{\tabcolsep}{1.5pt}
    \caption{Quantitative comparison between our approach and the current state-of-the-
art on CVGlobal street view to map retrieval.}
    \begin{tabular}{l|cccc|cccc|cccc}
    \toprule
        \multirow{2}{*}{Methods} & \multicolumn{4}{c|}{Regional map val} & \multicolumn{4}{c|}{Regional map test} & \multicolumn{4}{c}{} \\
         & R@1 & R@5 & R@10 & R@1\% & R@1 & R@5 & R@10 & R@1\% & & & & \\
        \midrule
        GeoDTR\cite{zhang2023cross} & 11.95 & 24.33 & 34.07 & 89.32 & 4.57 & 10.47 & 16.07 & 53.45 & & & & \\
        SAIG-D\cite{zhu2023simple} & 37.27 & 70.61 & 80.01 & 97.53 & 4.64 & 14.80 & 22.16 & 61.55 & & & & \\
        Samp4G\cite{Deuser_2023_ICCV} & 75.38 & 92.80 & 95.67 & 99.54 & 46.94 & 70.46 & 77.91 & 91.47 & & & & \\
        Ours & \textbf{81.31} & \textbf{95.84} & \textbf{97.68} & \textbf{99.71} & \textbf{49.41} & \textbf{71.04} & \textbf{78.33} & \textbf{91.96} & & & & \\
        \bottomrule
    \end{tabular}
    \label{tab:map_reginal}
\end{table}

\textbf{Cross-region image retrieval.} The experimental results from Table \ref{tab:reginal} show that the Regional-val, being a randomly divided validation set, provides extensive coverage of the training area, resulting in very ideal results. Meanwhile, the cross-regional evaluation performed on Regional-test, where the city style remains unchanged, demonstrates good algorithm performance. However, in the Cross-continent test, which faces scenes with significant style differences, performance noticeably declines. Compared to existing methods, our approach shows a clear improvement in performance, particularly standing out in challenging scenarios.

\textbf{Cross-temporal image retrieval.} As shown in Table \ref{tab:time}, we evaluate the model trained on 2023 data for cross-view assessment. We use ground data from 2013, 2016, and 2019, as well as a mix of these three years as queries, with 2023 satellite images as references. Our algorithm demonstrates superior performance in cross-temporal retrieval tasks. The model needs to capture long-term persistent features (e.g., road layouts and building structures) and reduce reliance on transient features (e.g., temporary buildings and vehicles) for this cross-temporal task. We also observe that the closer the data year is to 2023, the more ideal the evaluation results. Meanwhile, mixing data from different years introduces more disturbances to the results, but this scenario is closer to real-world application settings and represents a challenge that needs to be overcome.

\textbf{Street view to map retrieval.} We conducted training and cross-regional evaluations by replacing satellite images with map slice data, with results shown in Table \ref{tab:map_reginal}. Due to the lack of specific appearance and texture information in map data, the performance of various algorithms was not optimal. However, in the validation set, because of the high quality of map data used for both training and testing and the dense coverage of training data, the top-1 recall using map tiles exceeds 70\%, proving the effectiveness of applying map data in cross-view geolocalization tasks.

\section{Conclusion}
\label{sec:Conclusion}

In our work, we use the ground plane assumption and geometric relationships to convert street view panorama images to the satellite perspective, effectively reducing the domain gap between them. Our Panorama-BEV Co-Retrieval Network captures global information and enhances local detail features simultaneously. Our method achieves outstanding results on existing datasets. In the VIGOR-cross testing task, compared to the state-of-the-art methods, our top-1 recall rate increased by 10.49\%; in the CVUSA to CVACT task, it increased by 13.75\%. At the same time, we introduce a dataset that is closer to real-world scenarios as a benchmark. We also provide various evaluation modes to explore the performance of our method in cross-regional, cross-temporal, and map data retrieval tasks. Our proposed dataset offers a new testing platform for cross-view geographic localization, fostering new research in the field.

\begin{flushleft}
\textbf{Acknowledgements.} This project was funded by National Natural Science Foundation of China (Grant No. 42201358) and Shanghai AI Laboratory.
\end{flushleft}

\bibliographystyle{splncs04}
\bibliography{main}
\end{document}